\documentclass[runningheads]{llncs}
\usepackage[T1]{fontenc}
\usepackage{graphicx}
\usepackage{amsmath}
\usepackage{wrapfig}
\usepackage{xcolor}%
\usepackage{graphicx}
\usepackage{subfig} 

\usepackage{array}

\usepackage{amsmath}

\usepackage{multirow}

\usepackage[breaklinks=true,hidelinks]{hyperref}
\usepackage{breakcites}

\usepackage{subcaption}

\begin{document}
%




\title{Successful Misunderstandings: Learning to Coordinate Without Being Understood}


\titlerunning{Successful Misunderstandings}
%
\author{Nikolaos Kondylidis\inst{1}\orcidID{0000-0003-4304-564X} \and
Anil Yaman\inst{1}\orcidID{0000-0003-1379-3778} \and
Frank van Harmelen\inst{1}\orcidID{0000-0002-7913-0048} \and
Erman Acar\inst{2}\orcidID{0009-0002-9656-7249} \and
Annette ten Teije\inst{1}\orcidID{0000-0002-9771-8822}}
\authorrunning{N. Kondylidis et al.}

\institute{Vrije Universiteit Amsterdam, Amsterdam, the Netherlands
\email{\{nikos.kondylidis,a.yaman,frank.van.harmelen,annette.ten.teije\}@vu.nl}\\
\and
University of Amsterdam, Amsterdam, the Netherlands\\
\email{e.acar@uva.nl}}

\maketitle              
\begin{abstract}

The main approach to evaluating communication is by assessing how well it facilitates coordination. If two or more individuals can coordinate through communication, it is generally assumed that they understand one another.
We investigate this assumption in a signaling game where individuals develop a new vocabulary of signals to coordinate successfully.
In our game, the individuals do not have common observations besides the communication signal and outcome of the interaction, i.e. received reward.
This setting is used as a proxy to study communication emergence in populations of agents that perceive their environment very differently, e.g. hybrid populations that include humans and artificial agents.
%
%
%
%
%
%
%
Agents develop signals, use them, and refine interpretations while not observing how other agents are using them.
While populations always converge to optimal levels of coordination, in some cases, interacting agents interpret and use signals differently, converging to what we call \textbf{successful misunderstandings}.
%
%
However, agents of population that coordinate using misaligned interpretations, are unable to establish successful coordination with new interaction partners.
Not leading to coordination failure immediately, successful misunderstandings are difficult to spot and repair. 
Having at least three agents that all interact with each other are the two minimum conditions to ensure the emergence of shared interpretations.
Under these conditions, the agent population exhibits this emergent property of compensating for the lack of shared observations of signal use, ensuring the emergence of shared interpretations.


\keywords{Emergent Communication \and Successful Misunderstandings \and Emergent Communication from Disjoint Observations  \and Robust Emergent Semantics}
\end{abstract}

\section{Introduction}


The dynamic population aspect of Open Multi-Agent Systems (OMAS), does not guarantee an existing shared communication protocol.
Nonetheless, agents may need to coordinate to achieve their goals.
%
Furthermore, the system can exhibit robustness if it is possible to develop and maintain its communication system autonomously.
Emergent semantics \cite{emergent_semantics} describes how an agent population can develop a communication protocol in a decentralized manner from scratch.
Agents communicate signals in task-oriented interactions and individually update their interpretation based on interaction satisfaction.
Naturally, developing a shared semantics is expected to play a crucial role in aligning with other agents to achieve coordination.
However, the decentralized nature of emergent semantics makes it challenging to verify the correct or shared use of emergent semantics.
Successful coordination is usually assumed to suggest shared understandings or shared interpretations \cite{atencia_automata}, especially when the internal representations of the agents are inaccessible or uninterpretable \cite{shared_understanding_framework}.


Communication emergence and language alignment have been studied between artificial agents \cite{steels_language_2012,lazaridou,learning_to_communicate_forester,Lewis1970ConventionAP,aberer2003chatty} humans \cite{psychology_of_dialogue,human_communication_emergence}, and hybrid populations \cite{shared_understanding_framework,nevens2019interactive,human_agent_study}, although less in the last category.
Evaluating interpretations solely on the basis of the interaction outcome may foster new risks, as reward-driven optimization could lead to unintended behaviors \cite{russell2019human_compatible}, due to phenomena such as shortcut learning \cite{shortcut_learning}.
%
%
Shortcut learning describes scenarios where machine learning models achieve high task performance but for the wrong reasons, giving us the wrong impression that they are trained correctly while failing to generalize.
When it comes to language understanding, thought experiments, like the Chinese room \cite{Searle_1980_chinese_room} or the super intelligent octopus test \cite{bender_koller_octapus_test}, question whether humans and AI systems understand language in the same way.
Is it safe to assume successful coordination is always a consequence of shared interpretations for emergent semantics in hybrid populations?
%


In emergent semantics, signals are interpreted in relation to the (perceived) situational context of their use.
How are the emerged semantics affected in case the interacting agents have very different physiology and perception?
Answering this question can bring insights for when applying emergent semantic methods in hybrid populations that include both humans and artificial agents.
To computationally represent interaction between agents with very different perception, we take the extreme case where they exist in separate environments.
We study what interpretations emerge in such a setting and answer 3 research questions:
\vspace{-0.15cm}
\begin{itemize}
    \item \textbf{RQ1:} \textit{Is alignment always a by-product of coordination?}
    \item \textbf{RQ2:} \textit{When do misaligned interpretations cause coordination failure?}
    \item \textbf{RQ3:} \textit{What conditions ensure the emergence of shared interpretations?}
\end{itemize}
\vspace{-0.15cm}




%
%
%
We apply an emergent communication method, inspired by language games \cite{steels_language_2012}, to study the affect of interacting agents existing in separate environments.
%
An agent population develops a novel set of signals and aligns their interpretation by engaging in a Lewis signaling game \cite{Lewis1970ConventionAP} from separate environments.
%
During an interaction, a Sender agent needs to communicate the environment's observed state to a receiver, who can perform an action.
Existing in separate rooms, the Sender cannot observe the Receiver's action.
The agents always receive equal reward, the value of which depends on the environment's state and the performed action.
Which state-action combinations return positive reward is not given to the agents, otherwise they could infer unobserved information.
Agents reinforce (weaken) signal interpretations that lead to (un)successful interactions.
%




%
To answer RQ1, we propose metrics to measure signal interpretation alignment between agents and compare it with coordination success.
To answer RQ2, we design experiments that have two phases, i.e. i) the convergence phase and ii) the robustness test phase, to study the adaptability of converged semantics.
%
Agent populations converge to signal interpretations that always lead to successfully coordination during the convergence phase.
%
%
During the second phase, we allow agents to interact with new interaction partners, to test the robustness of the converged semantics.
To answer RQ3, we identify necessary conditions during the convergence phase to ensure the emergence of shared semantics.
We present in detail three experiments.
In the first two, populations of 2 and 3 agents are joined by a new agent during the robustness test phase.
In the third experiment, we apply interaction restrictions in populations of 3 agents during the convergence phase, which are lifted during the test phase.
%
%
%

Across all episodes, the agents converge to using signals in a way that always returns the best possible reward during the convergence phase of the experiments.
In some cases, agents coordinate successfully, i.e perform the action that gives them positive reward, without interpreting signals in the same way; answering RQ1.
%
%
When agents coordinate successfully while interpreting the communicated symbol differently, we say that it is due to ``\textbf{successful misunderstandings}''.
Communicating using successful misunderstanding does not cause coordination failure during the convergence phase.
However, agents of populations that coordinate while interpreting signals differently, are not able to establish successful coordination during test phase with agents they have not interacted with during the convergence phase; answering RQ2.
To ensure shared interpretation, we identify two conditions: i) the population has at least 3 agents, and ii) every agent can interact with any other agent; answering RQ3.
%


We show that agent coordination is not always a result of shared signal interpretation or use in emergent semantics between separate environments.
Instead, agents can rely on successful misunderstandings that are not robust to new agent interaction pairings.
Finally, we show an emergent property of populations that learn to communicate and include at least 3 agents that can all interact with each other.
Such populations can compensate for the lack of observability of the environment and avoid the emergence of successful misunderstandings.
This is crucial when agents assume shared interpretations to identify the reward function of the environment.
Avoiding successful misunderstandings can offer insights for future research on hybrid populations that can learn to communicate.

%
%
%
%

%
The paper is structured as follows.
In Sec. \ref{sec:related_work} we provide an overview of studies on (emergent) communication, showing that successful coordination is typically interpreted as evidence of shared understanding, identifying the literature gap.
In Sec. \ref{sec:experiment_discription} we present how agent populations interact and learn to interpret signals.
In Sec. \ref{sec:experimental_setup}, we describe the experiments that allow us to answer our RQs.
%
%
In Sec. \ref{sec:experiment_results}, we present experiment results and answer our RQs.
Finally, in Sec. \ref{sec:conclusion_and_future_work} we conclude our study and suggest future directions.

\section{Related Work}
\label{sec:related_work}
\vspace{-0.1cm}
In this section, we present related studies on (emergent) communication in populations of artificial agents, humans, and hybrid populations.
We identify the novelties of our study with respect to the related work.
%

\vspace{-0.1cm}
\subsection{Emergent Communication Between Artificial Agents}
\vspace{-0.1cm}
The study of emergent communication has been a focal point in multi-agent learning, where agents develop their own language-like protocols to facilitate coordination and task completion.
\cite{steels_language_2012} introduced the Naming Game, which shows how agents can self-organize vocabularies through reinforcement mechanisms.
The agents engage in episodes playing referential games, having to use a word to refer to objects based on properties of the object, like its color.
The agents operate using interpretable methods, demonstrating how the agents converge to a common vocabulary that the agents also interpret similarly.
\cite{lazaridou} studied communication emergence using deep learning methods in which the agents had to communicate which of the images presented was the correct.
Agents learn to interpret a communication vocabulary by relating it with visual characteristics.
The authors design the experimental setup in ways that would push the agents to relate the communicated symbols to known categories of the images.
The study presents agents' internal representations, but does not make conclusive statements regarding their correct or shared interpretation.
In \cite{atencia_automata}, and other Ontology Alignment (OA) studies \cite{jerome,anslow}, task-oriented interactions are used as a means of establishing a shared understanding between agents.
The fundamental hypothesis in \cite{atencia_automata} is that successful coordination suggests alignment, that is, shared interpretations.
However, most studies do not investigate whether agents use shared interpretations, as long as they coordinate successfully \cite{learning_to_communicate_forester,rita2022emergentcommunicationgeneralizationoverfitting}.
In this work, we study whether emergent semantics between agents that communicated between separate environments are shared.

\vspace{-0.1cm}
\subsection{(Emergent) Communication Between Humans}
\vspace{-0.1cm}
An overview study on how humans interpret dialogue and recover from misaligned interpretations (misunderstandings) is presented in \cite{psychology_of_dialogue}.
The overview suggests that conversation participants independently construct their own internal situational model that includes contextual information, useful for text production or comprehension.
Furthermore, the study claims that ``the linguistic representations that underlie coordinated dialogue come to be aligned''.
Translating the claim in computational terms, successful coordination suggests aligned internal representations and vice versa.
Moreover, misalignment of participants' situational models can lead to misunderstandings and unsuccessful dialogue.
In such a case, the participants align their situational models and reach a common ground through more dialogue.
%
In \cite{human_communication_emergence}, emergent semantics experiments are performed with humans.
During each experiment, two participants need to develop a new sign system to co-navigate over a partially observable map of rooms. 
They can only communicate using a modified seismograph, which requires them to develop a new set of signs and learn to interpret them in the same way to successfully coordinate.
The study focuses in particular on the type of signals that humans create and how they assign meaning to them.
The simple nature of the experiment allows for the interpretation of the new signs.
%
%
%
Both signal forms and interpretations are related to either i) enumerating the rooms, or ii) describing their coordinates, or iii) describing visual characteristics of the rooms.
By interpreting the emergent signs, we can also see that the participants learned to interpret the signs in the same way.
This acts as further evidence of the common assumption that successful coordination leads to shared interpretations and vice versa.
Our work investigates whether this assumption holds when the communicating parties perceive their environment very differently as if they exist in separate environments.

\vspace{-0.1cm}
\subsection{Emergent Communication in Hybrid Populations}
\vspace{-0.1cm}
The assumption that successful coordination is indirect evidence and the result of shared interpretations is extended also to hybrid populations \cite{shared_understanding_framework}
%
%
Following \cite{atencia_automata}, they suggest that agents in hybrid populations can align their interpretations through task-oriented interactions that require coordination.
This assumption is followed since in hybrid populations, agent's interpretations cannot be directly compared to evaluate similarity \cite{shared_understanding_framework}.
The same assumption is held in a demo \cite{nevens2019interactive}, where humans can teach agents a grounded concept by playing the Naming Game.
In \cite{human_agent_study}, a human is teaching a robot to perform a collection of commands.
Again, successful task performance is interpreted as correct task understanding.
In this work, we investigate if this is indeed the case, when the interacting agents exist in separate environments, to computationally mimic interacting agents with very different perceptions.

\vspace{-0.1cm}
\subsection{Conclusion}
\vspace{-0.1cm}
All emergent semantics studies between artificial agents, humans, or hybrid populations follow the assumption that successful coordination suggests aligned representations.
Our RQ1, i.e. ``is alignment always a by-product of coordination?'' investigates whether this claim is always true, or if consistent coordination can also be observed with misaligned semantics.
%
To the best of our knowledge, this has not been studied before.
%
In case RQ1 shows that misalignment can still lead to consistent coordination, we then explore RQ2 and RQ3 that study what are the limitations in this case and how to prevent this from happening, respectively.

\vspace{-0.1cm}

\section{Emergent Communication over Separate Environments}
\label{sec:experiment_discription}
%
\vspace{-0.1cm}
%
We perform communication emergence experiments on a population of artificial agents, following the experimental setup of Language Games \cite{steels_language_2012}.
In Language Games, an agent population develops a new signaling system through pairwise participation in coordination games.
In contrast to Language Games, the agents do not interact in the same environment, and have very few common observations.
%
In addition to studying successful coordination, we are also recording to what extent the agents interpret the communication signals in the same way.

\begin{figure}[h!]
\centering
\includegraphics[width=\textwidth]{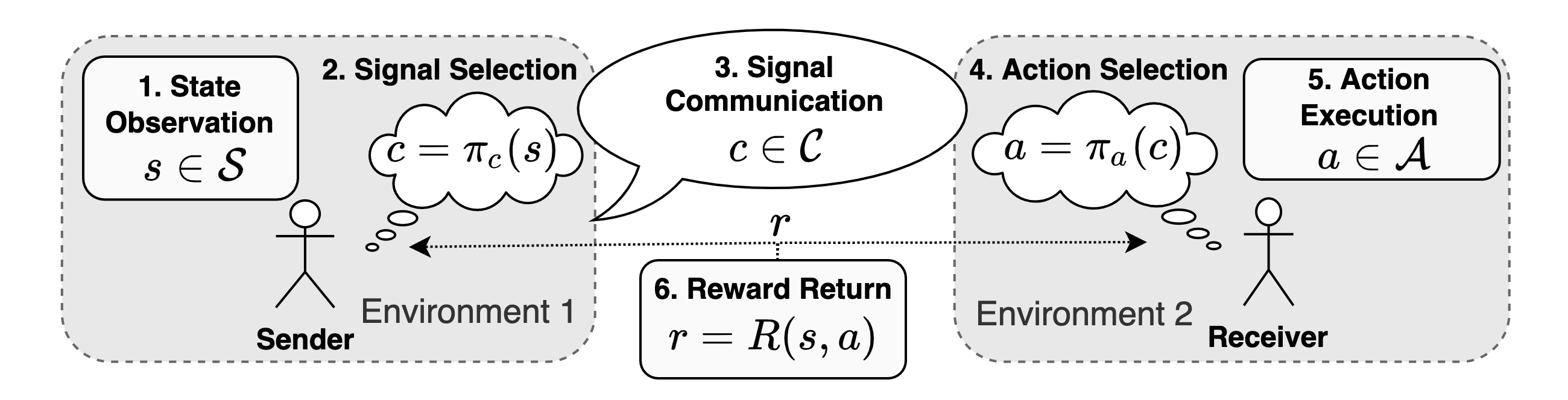}
\caption{\label{fig:Separate_Rooms} 
The 6 steps of the signaling game between a Sender and Receiver that exist in separate environments.
}
\end{figure}

\vspace{-0.1cm}

\subsection{The Signaling Game Over Separate Environments}
\vspace{-0.1cm}
This section introduces the Lewis signaling game \cite{Lewis1970ConventionAP} in which two selected agents communicate within an episode to coordinate.
One agent is assigned to the role of Sender, and the other to the role of Receiver.
The game takes place in an environment, that is, in some state $s \in \mathcal{S}$.
A set of actions ($\mathcal{A}$) can be performed in the environment.
Depending on the combination of the environment state $s$, and the action performed ($a \in \mathcal{A}$), a reward $r \in \mathcal{R}$ is given to agents, based on the reward function $R:\mathcal{S}\times \mathcal{A} \rightarrow \mathcal{R}$.
%
%
The reward function $R(s,a)$ is designed so that for each environment state $s$, there is exactly one action $a$ that returns a positive reward.
The agents always receive the same reward, making it a game of perfect common interest.

In our setting, the two agents play the coordination game from separate environments, as illustrated in Fig. \ref{fig:Separate_Rooms}.
At the beginning of the episode, the environment's state ($s$) is randomly assigned and can only be observed by the Sender.
Then, it decides which signal ($c$) to communicate, using its communication policy $\pi_c(s):\mathcal{S} \rightarrow \mathcal{C}$.
%
%
The Receiver is the only one that can select and perform an action ($a$).
To do so, it uses its action selection policy $\pi_a(c): \mathcal{C} \rightarrow \mathcal{A}$.
The Sender does not know which actions are performed by the Receiver.
Given the combination of state-action, the environment provides a reward value to the agents based on the reward function $R(s,a)$.
The agents do not know the reward function of the environment $R(s,a)$.
The two possible reward functions for an environment with two possible states and two actions is presented in Fig. \ref{fig:Game_A_and_B}.
Finally the Sender and the Receiver update how they use the communicated signal based on the interaction outcome.
%
Except for the signal communicated and the reward received, the agents have no other common observations.

\vspace{-0.1cm}
\subsection{Agent Policies}
\vspace{-0.1cm}
This section describes signal and action selection policies, and how agents learn from episodes.

\subsubsection{Signal Utility Values \& Interpretations}
Signals are used probabilistically, according to the total reward each interpretation has returned in the past.
Agents are only aware of episodes they have participated in and only learn from them.
For each signal, the agent stores the total reward, indexed by observed state $V(c,s)$ or performed action $V(c,a)$, depending on their role in that episode.
Agents forget words they did not use in the last 20 episodes in which they participated.

\vspace{-0.1cm}
\subsubsection{Signal Selection Policy $\pi_c(s)$}
Function $\pi_c(s)$ takes as input the observed state $s$ and decides which signal $c$ to communicate.
First, the agent populates a set of candidate signals to communicate $\mathcal{C}_s$; its action set.
The action set consists of all signals that have a higher utility value for the observed state than any other state; $c \in \mathcal{C}_s$, if $V(c,s) > V(c, s^\prime) \in S \setminus \{s\})$.
The utility values of the candidate signals are normalized using the softmax function and treated as selection probabilities for the corresponding signals.
To ensure that the agents explore enough and eventually converge to using an optimal policy, a probability mass equal to $\epsilon$ is evenly distributed among the candidate words.
The value of $\epsilon$ decreases linearly from 0.2 to 0 during the first 5k episodes of the experiment.
In case there are no candidate signals to communicate, i.e. $\mathcal{C}_s = \emptyset$, the agent creates and communicates a new signal.

\vspace{-0.1cm}
\subsubsection{Action Selection Policy $\pi_a(c)$}

The Receiver observes the communicated signal $c$ and retrieves the total returns per action this agent has received in the past when this word was sent, $V_r(c,a), \forall a \in \mathcal{A}$.
Then, it normalizes these values using the softmax function and probabilistically selects one of them.
Again, an $\epsilon$ value or probability mass is assigned to all possible actions $a \in \mathcal{A}$.
%
Again, the value of $\epsilon$ decreases linearly from 0.2 to 0 during the first 5k episodes of the experiment.

\vspace{-0.1cm}
\subsubsection{Learning from Episodes}

At the end of the episode, the agents receive some reward based on the combination of the environment's state and the action performed by the Receiver, as defined by $r = R(s,a) \in \mathcal{R}$.
The Sender updates the utility of the sent signal for the observed state, $V_s(c,s) = V_s(c,s) + r $.
The Receiver does the same for the utility of the received signal and the performed action, $V_r(c,a) = V_r(c,a) + r $.


\vspace{-0.1cm}

\section{Experimental Setup}
\label{sec:experimental_setup}

\vspace{-0.1cm}


In this section, we describe the experimental setup that allows us to answer our 3 RQs.
In Sec. \ref{subsec:eval_metrics}, we introduce evaluation metrics that allow us to measure both coordination success and interpretation alignment in a population.
To answer RQ1, whether interpretations that enable successful coordination are always shared, we just need one example where agents coordinate successfully with misaligned interpretations.
%
To answer RQ2, whether useful but misaligned representations can cause coordination failure in the future, we perform experiments in two phases, described in Sec. \ref{subsec:two_stages_explanation}.
During the first phase of the experiment, i.e. the convergence phase, the population converges to using a common set of signals that allow them to achieve optimal coordination levels.
During the second phase, i.e. the robustness test phase, we allow the agents to also interact with new interaction partners.
Finally, in \ref{subsec:exp_setup_overview}, we present an overview of the 3 experiments we perform and their conditions, which allows us to answer RQ3, i.e. what are the necessary conditions to ensure the emergence of shared interpretations?

%


%
%


\vspace{-0.1cm}

\subsection{Evaluation Metrics}
\label{subsec:eval_metrics}

\vspace{-0.1cm}

We will now present the metrics that allow us to evaluate the agents' emerged semantics.
We report average metric values over moving windows of 100 episodes for the findings to be more robust.

\begin{itemize}
    \item \textbf{Reward}: The average reward reflects the ability of the agent population to coordinate successfully.
    \item \textbf{|Vocabulary|}: The agents' average signal vocabulary size. Reduction shows population semantics convergence. 

    \item \textbf{Alignment}: Semantic alignment measures the ratio of the population that shares the dominant interpretation for each signal-role (Sender or Receiver) combination.
    It is calculated by the formula $\frac{|interpretation\_ majority| - 1} {|population| - 1}$, and averaged over all signal-role combinations.
    The optimal value is 1, when every agent uses all signals in the same way for every roles.


%
    
    
    \item \textbf{Intent Met}: The ratio of episodes in which the Sender's intent was met, i.e. the Sender would select the same action as the receiver did, given the communicated signal.
    \item \textbf{Suc. Mis.}: The successful misunderstanding ratio of episodes, when agents receive positive reward while the Sender's intent was not met.
\end{itemize}

\subsection{The 2-Phase Robustness Tests}
\label{subsec:two_stages_explanation}
To answer our second research question, i.e. are there negative consequences if the interactions are successful using misaligned representations, we perform experiments in two phases.
In \textbf{Phase 1}, i.e., convergence phase, agents with no prior knowledge are interacting until they converge using signals that allow them to always coordinate successfully.
Then, in \textbf{Phase 2}, i.e. robustness test phase, agents interact with new interaction partners and see if after enough interactions, they are able to converge again to always coordinating successfully.
Each phase consists of 10k episodes, giving agents enough interactions to converge.
We have two ways to make agents interact with new interaction partners in phase two, which gives us two experimental setups.
\begin{itemize}
    \item \textbf{Population Increase}: Agents can interact with any other agent. A new agent with no knowledge is added to the population at the start of phase 2.
    \item \textbf{Initially Grouped}: In phase 1, agents are split into two groups and can only interact with agents from the other group. In phase 2, everyone interacts with everyone.
\end{itemize}


\subsection{Overview of Presented Experiments}
\label{subsec:exp_setup_overview}
\vspace{-0.1cm}

In this section, we describe the three experiments that we present in detail and allow us to answer our RQs.
The first experiment starts with 2 agents in the first phase; named \textbf{ 2 agents}.
In the second phase, we perform the population increase robustness test, by adding a third agent.
%
%
Next, we conduct an experiment with three agents that form interaction pairs in an unrestricted way during the first phase; denoted as \textbf{3 Unrestricted}.
During its test phase, we again apply the new population increase robustness test, adding a fourth agent.
The third experiment, that is, \textbf{3 Restricted}, starts with 3 agents, but two of them belong to the same group and are not allowed to interact during the first phase.
Thus, this experiment applies the ``initially grouped'' robustness test, and during the second phase, the two agents of the same group are also allowed to interact. 
This experiment is repeated with 4 agents in both phases, to further validate our observations.
Tab. \ref{tab:exp_setup_overview} provides an overview of these 3 experiments.

%


\begin{table}[h]
\centering
\caption{ Experimental setup overview of the three presented experiments.\label{tab:exp_setup_overview}}
\begin{tabular}{|l|c|c|c|c|c|c|}
\hline
\multicolumn{2}{|c|}{Experiments } & \multicolumn{2}{c|}{Phase 1: Convergence} & Intervention & \multicolumn{2}{c|}{Phase 2: Robust. Test} \\ \hline
 \#   &     Name                   & Restricted          & Agents           & Robustness Test         & Restricted            & Agents            \\ \hline
1& \textbf{2 Agents}                        & No (N/A)            & 2                & Population Increase              & No                    & 3                 \\
2& \textbf{3 Unrestricted}                  & No                  & 3                & Population Increase              & No                    & 4                 \\
3& \textbf{3 Restricted}                    & Yes                 & 3                & Initially Grouped                & No                    & 3          \\      \hline
\end{tabular}
\end{table}

\begin{figure}[h!]
\centering
\includegraphics[width=0.7\textwidth]{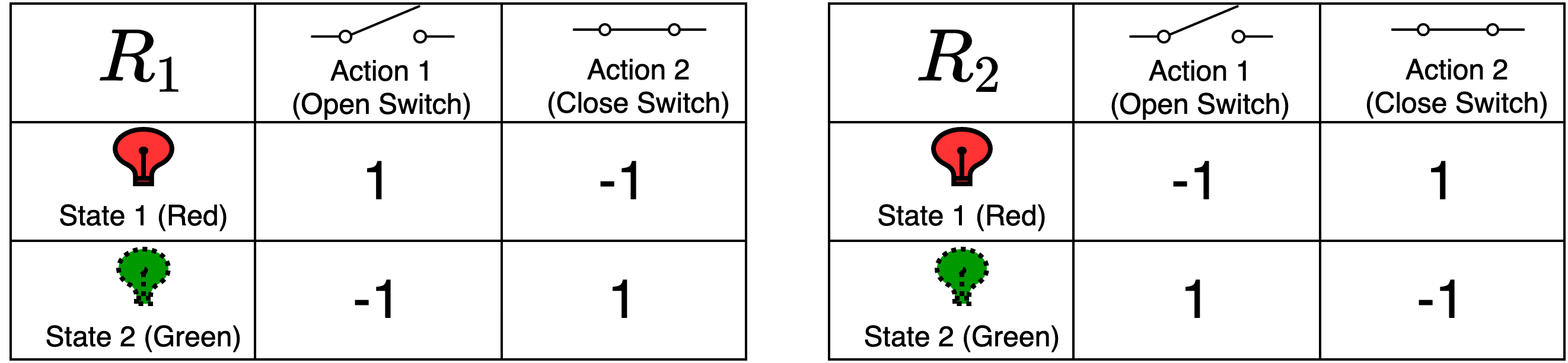}
\caption{ \label{fig:Game_A_and_B} The two possible reward functions ($R_1, R_2$) for a setting with 2 states and 2 actions. \vspace{-0.2cm}
}
\end{figure}

%
Experiments are performed using environments with 2 states and 2 actions.
Consequently, the only two possible reward functions are $R_1$ and $R_2$ for $\mathcal{R} = \{-1, 1\}$, which are presented in Fig. \ref{fig:Game_A_and_B}.
To minimize the effect of random chance in the presented experiments, we repeat them 1000 times and report average values together with their Standard Deviation (SD) in parentheses.
Each repetition of the experiment independently samples between using $R_1$ or $R_2$.
%
%
The code for replicating the findings is available online\footnote[1]{\url{https://github.com/kondilidisn/eumas-successful-misunderstandings}}.

\vspace{-0.2cm}

\section{Experiment Results}
\label{sec:experiment_results}
\vspace{-0.2cm}

In this section, we present the experiment results, perform necessary analysis, and answer our 3 RQs.
Experiments 1, 2, and 3, as described in Sec. \ref{subsec:exp_setup_overview}, are presented in Sec. \ref{subsec:pop_increase_exp_2_to_3}, \ref{subsec:pop_increase_exp_3_to_4}, and \ref{subsec:pairing_restrictions_exp}, respectively.
In Sec. \ref{subsec:necessary_conditions_summary} we identify the necessary conditions that ensure shared signal interpretations in emergent communication over separate environments.
In Sec. \ref{subsec:validation_on_more_complex_reward_function}, we mention performed experiments using more complicated reward functions that validate the generalizability of our findings.

\vspace{-0.3cm}


\subsection{Experiment 1: 2 Agents}
\label{subsec:pop_increase_exp_2_to_3}
\begin{figure}[h]

\includegraphics[width=\textwidth]{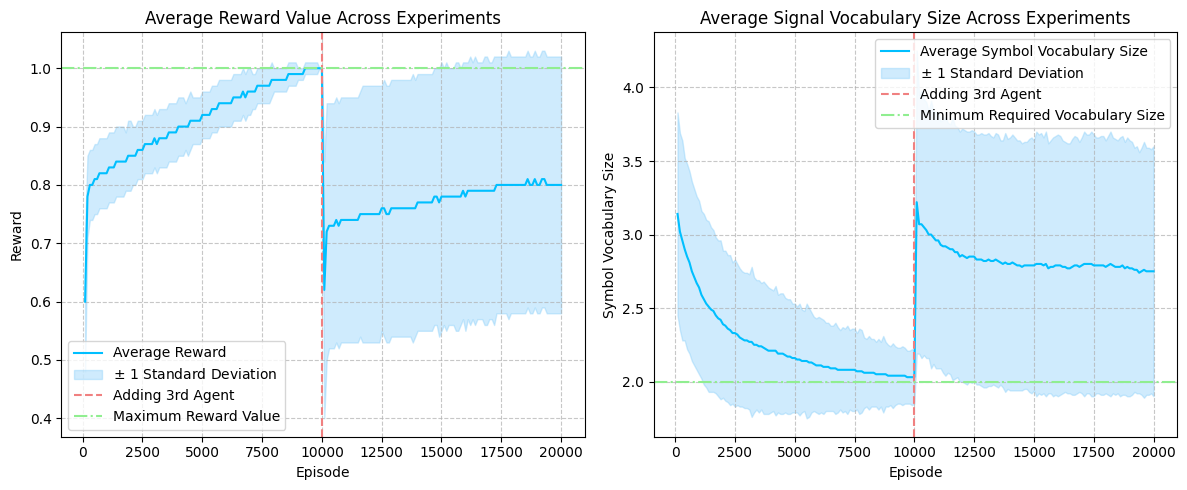}
    \caption{Average reward (left) and signal vocabulary size (right) of the 2 Agents experiment.} 
    \label{fig:2_agents_become_3_convergence_success_and_failure_plots_new}
    
\end{figure}

\vspace{-0.1cm}
Here, we will present the results of our first experiment.
During the first phase, i.e. until episode 10k, populations of 2 agents exhibit clear signs of communication emergence convergence.
As shown in Fig. \ref{fig:2_agents_become_3_convergence_success_and_failure_plots_new}, agents receive maximum reward while requiring approximately the least necessary number of signals.
%
%
At the same time, i) semantic alignment score, ii) intent met ratio, and iii) successful misunderstandings ratio are on average around 50\% at the end of the first phase, as Tab. \ref{tab:2_agents_become_3} shows.
This means that on average half the times i) the agents interpret the signals differently, ii) the Sender would behave differently had it acting as Receiver, and iii) the agents receive high reward for the wrong reasons.
This answers our first research question, proving that emergent semantics that always lead to successful coordination do not always imply shared interpretations.
Once a third agent is introduced in the test phase, the population cannot converge again to optimal behavior.
Instead, it only manages to achieve 80\% of the maximum reward, using more words than necessary.
On the other hand, semantic alignment and Sender's intent met ratios have increased to 0.8 and 0.7, respectively.

\begin{table}[h] 
\centering
\captionsetup{position=top} 
\caption{\label{tab:2_agents_become_3}
Evolution of the evaluation metric values over episodes of both phases, for the 1st experiment, i.e. ``2 Agents''.
SD in parenthesis.
}
{

\begin{tabular}{ >{\centering\arraybackslash}p{0.13\textwidth} | >{\centering\arraybackslash}p{0.15\textwidth}| >{\centering\arraybackslash}p{0.18\textwidth} | >{\centering\arraybackslash}p{0.15\textwidth} | >{\centering\arraybackslash}p{0.18\textwidth} | >{\centering\arraybackslash}p{0.15\textwidth} } 
\textbf{Episodes} & \textbf{Reward} & \textbf{|Vocabulary|} & \textbf{Alignment} & \textbf{Intent Met} & \textbf{Suc. Mis.} \\ \hline
\multicolumn{6}{c}{ \textbf{Convergence Phase}: 2 agents }  \\ \hline

 100 &   0.60 (0.12) &   3.14 (0.69) &   0.49 (0.29) &   0.49 (0.15) &   0.51 (0.15) \\ 
 500 &   0.81 (0.06) &   2.85 (0.63) &   0.48 (0.34) &   0.49 (0.23) &   0.51 (0.23) \\ 
 5k &   0.92 (0.04) &   2.16 (0.37) &   0.48 (0.46) &   0.49 (0.39) &   0.51 (0.39) \\ 
 10k &   1.00 (0.00) &   2.03 (0.18) &   0.48 (0.49) &   0.49 (0.49) &   0.51 (0.49) \\  \hline

\multicolumn{6}{c}{ \textbf{Robustness Test Phase}: 3 Agents (Unrestricted Partners)}  \\ \hline

10k+100 &   0.62 (0.22) &   3.22 (1.03) &   0.73 (0.21) &   0.59 (0.27) &   0.41 (0.27) \\ 
10k+500 &   0.73 (0.21) &   3.03 (0.87) &   0.76 (0.21) &   0.62 (0.29) &   0.38 (0.29) \\ 
10k+5k &   0.78 (0.23) &   2.79 (0.85) &   0.79 (0.21) &   0.67 (0.31) &   0.33 (0.31) \\ 
10k+10k &   0.80 (0.22) &   2.75 (0.85) &   0.80 (0.21) &   0.70 (0.31) &   0.30 (0.31) \\

\end{tabular}
}

\end{table}

\begin{figure}[h!]

\includegraphics[width=\textwidth]{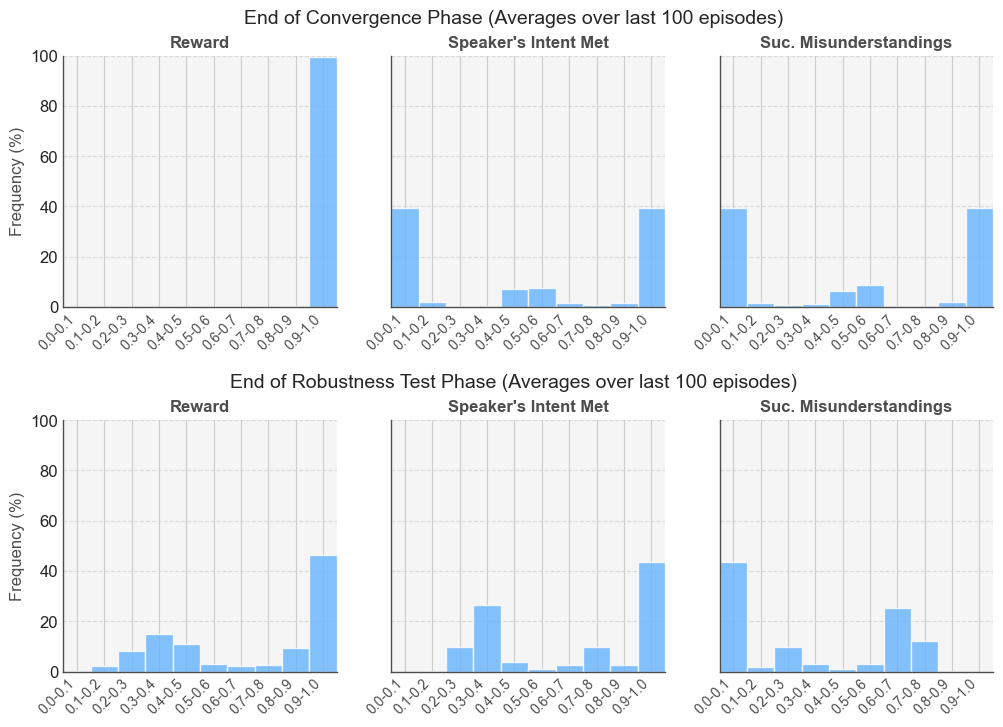}

    \caption{Distribution of the individual experiments over value ranges for 3 evaluation metrics at the end of first (top) and second (bottom) phases of the 1st experiment, i.e. ``2 Agents''.}
    \label{fig:2_agents_converging_behavior_unable_to_recover_histograms}
\end{figure}

\vspace{-0.5cm}

\subsubsection{Failure Analysis}
There are two main questions raised from the experiment behavior presented in Tab. \ref{tab:2_agents_become_3}.
First, how do agents coordinate successfully while misinterpreting each other during the convergence phase?
Second, why do populations cannot converge to successfully coordinating during the second phase, when the third agent is introduced?
%
%
To answer them, we present histograms of average evaluation metrics of individual experiments at the end of the two phases, i.e. episodes 10k and 20k, in Fig. \ref{fig:2_agents_converging_behavior_unable_to_recover_histograms}.
%
%
At the end of the convergence phase, Sender's intent was always met (0.9-1.0) at 40\% of the experiments.
For another 40\% of the experiments, Sender's intent was never met (0.0-0.1).
However, the agents were always communicating successfully.
The 40\% of the experiments where Sender's intent met was (>0.9) is still present at the end of the second phase.
This demonstrates that populations with shared interpretations manage to reconverge to successfully coordinating after the fourth agent is introduced in the second phase.
Instead, 40\% of the experiments in which the agents converge to using successful misunderstandings in the first phase ((>0.9)) are scattered both in the Sender's intent met and in the reward histograms around random chance (0.5).
This shows that agents of populations that converge to using misaligned interpretations are unable to adapt to interacting with new agents; responding to RQ2.


%

\begin{figure}[h!]
\centering
\includegraphics[width=\textwidth]{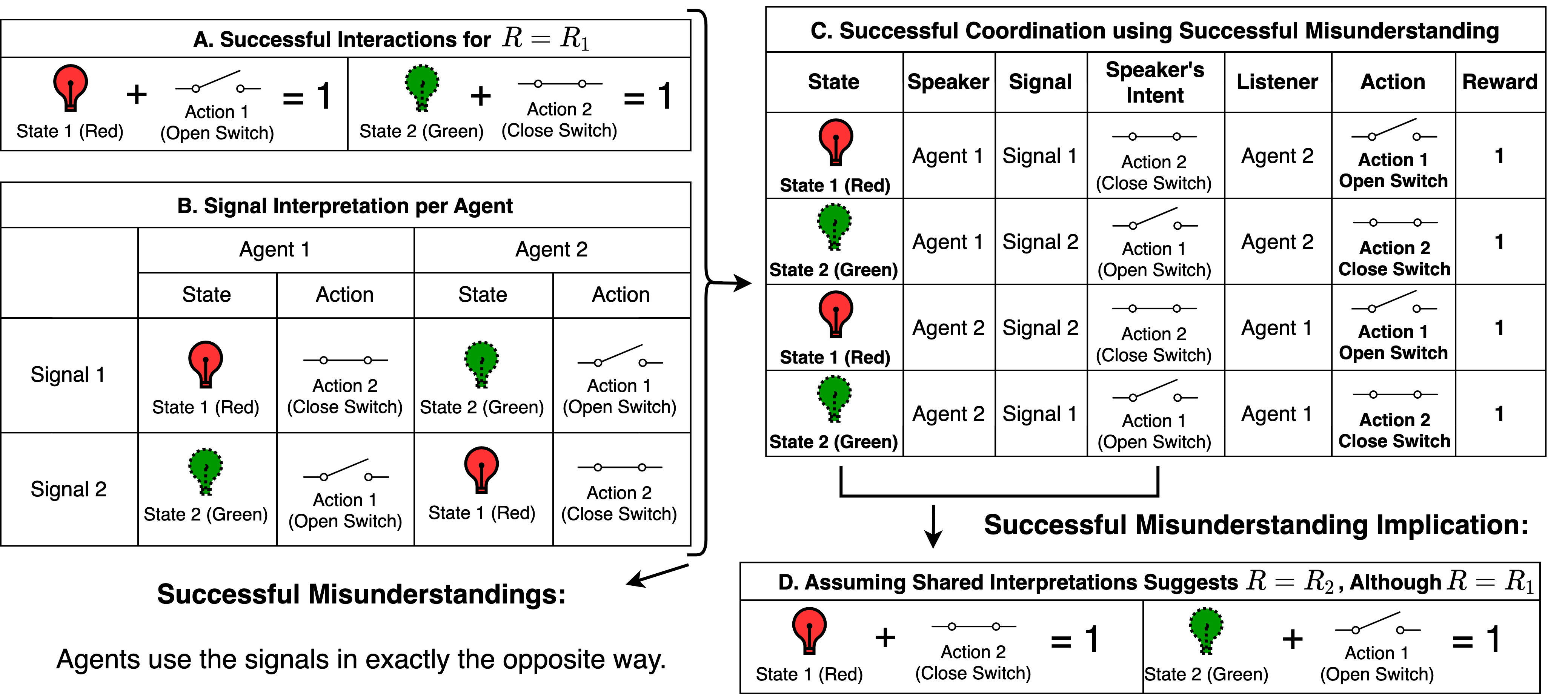}
\caption{
\label{fig:succ_mis_analysis}
A: Experiment reward function.
B: Agent signal interpretation.
C: Successful coordination due to successful misunderstandings in all 4 possible episodes.
D: Inferred reward function, assuming common interpretations.
}
\end{figure}

%
By investigating the performance of individual experiments in Fig. \ref{fig:2_agents_converging_behavior_unable_to_recover_histograms}, we observe that our experimental setup has two Nash equilibria.
One, is the ''expected one'', when the agents interpret the signals in the same way, which is always the case for 40\% of the experiments presented above.
The second is when the agents use the signals exactly in the opposite way, which is the case for the other 40\% of the experiments.
When the agents converge to the second equilibrium, we say that they coordinate correctly using \textbf{successful misunderstandings}.
Fig. \ref{fig:succ_mis_analysis} explains this in detail.
%
%
%

\subsection{Experiment 2: 3 Unrestricted agents}
\label{subsec:pop_increase_exp_3_to_4}

\begin{figure}[h]

\includegraphics[width=\textwidth]{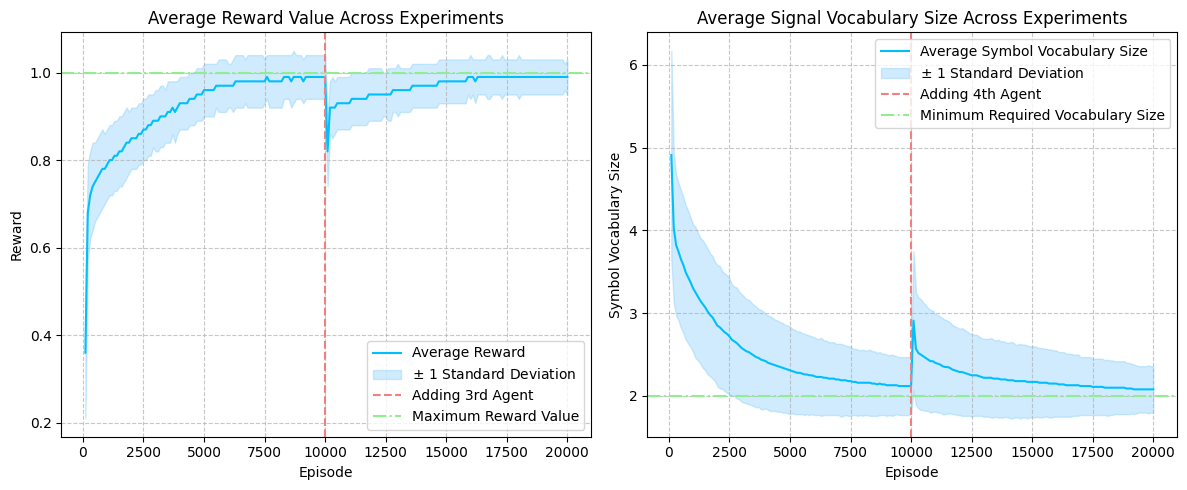}
    \caption{Experiment ``3 Unrestricted'' showing successful communication convergence and passing the robustness test.
    }     \label{fig:3_agents_become_4_random_interactions_convergence_success}
\end{figure}

We will now present the results of the second experiment, where the population consists of 3 agents from the beginning.
%
%
%
%
An initial population of 3 agents approximates optimal communication behavior in terms of reward and vocabulary needed, at the end of both experiment phases, as Fig. \ref{fig:3_agents_become_4_random_interactions_convergence_success} illustrates.
Furthermore, populations are able to converge again to optimal behavior when a new agent is added at the beginning of phase 2 (red dashed lines), passing the robustness test.
Agents almost always use communication signals in the same way, since both the alignment score and the Sender's intent met ratio are around 100\%, as Tab. \ref{tab:3_agents_solution} shows.


%

%
%


\begin{table}[h] 
\centering
\captionsetup{position=top}
\caption{\label{tab:3_agents_solution}
Evolution of the evaluation metric values over episodes of both phases, for the 2nd experiment, i.e. ``3 Unrestricted''.
SD in parenthesis.
}
{
\begin{tabular}{ >{\centering\arraybackslash}p{0.13\textwidth} | >{\centering\arraybackslash}p{0.15\textwidth}| >{\centering\arraybackslash}p{0.18\textwidth} | >{\centering\arraybackslash}p{0.15\textwidth} | >{\centering\arraybackslash}p{0.18\textwidth} | >{\centering\arraybackslash}p{0.15\textwidth} } 
\textbf{Episodes} & \textbf{Reward} & \textbf{|Vocabulary|} & \textbf{Alignment} & \textbf{Intent Met} & \textbf{Suc. Mis.} \\ \hline
\multicolumn{6}{c}{ \textbf{Convergence Phase}: 3 Agents (Unrestricted Partners)}  \\ \hline
100 &   0.36 (0.15) &   4.91 (1.26) &   0.70 (0.13) &   0.56 (0.08) &   0.44 (0.08) \\ 
 500 &   0.75 (0.09) &   3.65 (0.83) &   0.86 (0.11) &   0.73 (0.11) &   0.27 (0.11) \\ 
 5k &   0.96 (0.06) &   2.31 (0.53) &   0.96 (0.08) &   0.94 (0.09) &   0.06 (0.09) \\ 
 10k &   0.99 (0.05) &   2.12 (0.34) &   0.98 (0.07) &   0.98 (0.08) &   0.02 (0.08) \\ \hline
\multicolumn{6}{c}{ \textbf{Robustness Test Phase}: 4 Agents (Unrestricted Partners)}  \\ \hline
10k + 100 &   0.82 (0.08) &   2.91 (0.84) &   0.86 (0.14) &   0.88 (0.07) &   0.12 (0.07) \\ 
10k + 500 &   0.93 (0.06) &   2.48 (0.65) &   0.94 (0.09) &   0.92 (0.06) &   0.08 (0.06) \\ 
10k + 5k &   0.98 (0.05) &   2.17 (0.43) &   0.98 (0.06) &   0.97 (0.05) &   0.03 (0.05) \\ 
10k + 10k &   0.99 (0.04) &   2.08 (0.28) &   0.99 (0.05) &   0.99 (0.04) &   0.01 (0.04) \\
\end{tabular}
}
\end{table}

\newpage

\subsection{Experiment 3: 3 Restricted agents} 
\label{subsec:pairing_restrictions_exp}

\vspace{-0.1cm}

We present the results of the third experiment in Tab. \ref{tab:3_agents_biased_interactions}, where interaction restrictions are applied to a population of 3 agents during phase 1.
The emergent communication behavior of the experiment is similar to the one seen in the first experiment; see Sec. \ref{subsec:pop_increase_exp_2_to_3}.
Specifically, during phase 1, agents are able to coordinate successfully.
At the same time, populations have an average semantic alignment score of 68\%, i.e. equal to random change for 3 agent populations, and Sender's intent is only met half the times.
This again causes the experiment to fail the robustness test during phase 2, achieving on average 0.71 reward by the end of the second phase, using 2.24 signals; see Tab. \ref{tab:3_agents_biased_interactions}.

\begin{table}[htbp] 
\centering
\captionsetup{position=top} 

\caption{\label{tab:3_agents_biased_interactions}
Evolution of the evaluation metric values over episodes of both phases, for the 3rd experiment, i.e. ``3 Restricted''.
SD in parenthesis.}
{ \small

\begin{tabular}{ >{\centering\arraybackslash}p{0.13\textwidth} | >{\centering\arraybackslash}p{0.15\textwidth}| >{\centering\arraybackslash}p{0.18\textwidth} | >{\centering\arraybackslash}p{0.15\textwidth} | >{\centering\arraybackslash}p{0.18\textwidth} | >{\centering\arraybackslash}p{0.15\textwidth} } 
\textbf{Episodes} & \textbf{Reward} & \textbf{|Vocabulary|} & \textbf{Alignment} & \textbf{Intent Met} & \textbf{Suc. Mis.} \\ \hline
\multicolumn{6}{c}{ \textbf{Convergence Phase}: 3 Agents (Restricted Partners)}  \\ \hline

100 &   0.46 (0.14) &   4.98 (1.41) &   0.54 (0.15) &   0.50 (0.12) &   0.50 (0.12) \\ 
 500 &   0.79 (0.07) &   3.72 (0.83) &   0.62 (0.17) &   0.50 (0.21) &   0.50 (0.21) \\ 
 5k &   0.95 (0.03) &   2.75 (0.89) &   0.65 (0.23) &   0.50 (0.38) &   0.50 (0.38) \\ 
 10k &   0.99 (0.02) &   2.48 (0.82) &   0.68 (0.25) &   0.50 (0.45) &   0.50 (0.45) \\ \hline

\multicolumn{6}{c}{ \textbf{Robustness Test Phase}: 3 Agents (Unrestricted Partners)}  \\ \hline

10k + 100 &   0.69 (0.30) &   2.47 (0.81) &   0.74 (0.22) &   0.65 (0.30) &   0.35 (0.30) \\ 
10k + 500 &   0.70 (0.30) &   2.44 (0.76) &   0.74 (0.22) &   0.65 (0.30) &   0.35 (0.30) \\ 
10k + 5k &   0.71 (0.31) &   2.27 (0.53) &   0.75 (0.23) &   0.68 (0.31) &   0.32 (0.31) \\ 
10k + 10k &   0.71 (0.31) &   2.24 (0.49) &   0.76 (0.23) &   0.69 (0.31) &   0.31 (0.31) \\

\end{tabular}
}
\end{table}

\vspace{-0.1cm}

\paragraph{Beyond Three Agents: Consistent Findings with a Fourth.}
%
We repeat the experiment with 4 agents (in both phases) to see if a larger population size is a sufficient condition for avoiding successful misunderstandings.
%
Nevertheless, the findings are consistent with those presented in Tab. \ref{tab:3_agents_biased_interactions}.
%
Due to space limitation, we will only present key findings.
During the convergence phase, all experiments converge to optimal communication, receiving maximum reward with the minimum required signal vocabulary size.
However, in approximately half of the experiments, this is achieved using successful misunderstandings.
%
During the second phase, when agent pairing restrictions are lifted, the populations do not converge to optimal levels of coordination.
By the end of the second phase, the populations on average receive 0.68 (0.29) reward, uses 2.54 (0.68) signals, has a semantic alignment ratio of 0.68 (0.29), and the sender's intent is met 0.68\% (29\%) of the times (standard deviation in parentheses).
Interestingly, we observe that at the end of the first phase, agents that belong to the same group (who are not yet allowed to interact) use signals in a more similar way than when comparing agents from different groups.
%
Specifically, each group of two agents has an average semantic alignment score of 0.86 (0.20), while the total population has a semantic alignment score of 0.64 (0.25).

\vspace{-0.2cm}

\subsection{Necessary Conditions to Avoid Successful Misunderstandings}
\label{subsec:necessary_conditions_summary}
\vspace{-0.1cm}

Experiments ``2 Agents'' and ``3 Restricted'' allow the population to converge using successful misunderstandings during phase 1.
On the other hand, experiment ``3 Unrestricted'' does not. 
%
We can therefore conclude that there are two necessary conditions for agents to develop shared interpretations when learning to communicate from separate environments and do not know the environment's reward function.
First, the population must consist of at least 3 agents.
Second, agents must be expected to interact with any other agent.
%
In case we want to ensure shared emerged semantics between only two agents, we must initially include a third agent, who can later be removed.
For larger populations, it is sufficient to have a ``misunderstanding detection'' team of 3 agents. 
When these 3 agents interact successfully with each other for any scenario, then they have shared interpretations.
%
%
Subsequently, any agent that interacts successfully with one of these 3 agents, can be assumed to also share their interpretations.




\vspace{-0.13cm}

\subsection{Experiment Validation on More Complex Reward Functions}
\label{subsec:validation_on_more_complex_reward_function}
\vspace{-0.1cm}

\begin{figure}[h!]
\vspace*{-0.15cm}
\centering
\includegraphics[width=0.5\textwidth]{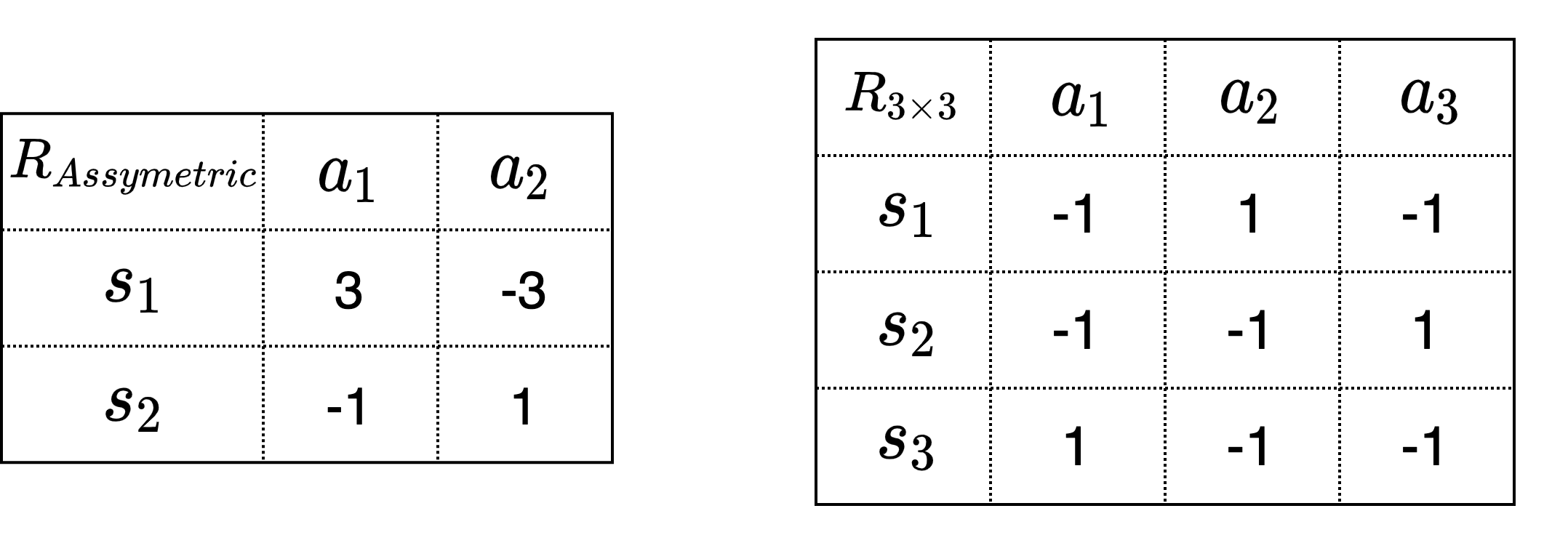}
\vspace*{-0.3cm}
\caption{ \vspace{-0.15cm} \label{fig:games_c_and_d} Other reward functions used to validate our findings' generalisability.}
\end{figure}

To validate the generalizability of our findings, we repeat our experiments for two additional settings that make coordination establishment more challenging.
We present a possible reward function for each new setting in Fig. \ref{fig:games_c_and_d}.
In the first setting, i.e. $R_{Assymetric}$, reward functions are asymmetric, biasing the action selection of the receiver and making coordination establishment more challenging. 
%
%
%
%
In the second setting, i.e. $R_{3\times3}$, the environment has 3 states and allows 3 actions, as opposed to 2 states and 2 actions used before.
This leaves more room for possible misinterpretations and decreases the possibility of successful coordination due to chance.
%
These additional experiments verify that having at least three agents and ensuring a fair chance that every agent interacts with any other agent are the two necessary conditions for shared emerged interpretations.



\vspace{-0.3cm}

\section{Conclusions and Future Work}
\label{sec:conclusion_and_future_work}



In this work, we extend emergent semantics experiments to interactions that take place between separate environments.
The experiments presented allow us to answer our 3 RQs.
By comparing measurements of coordination success and signal interpretation alignment between agents, we are able to answer RQ1, i.e. ``Is alignment always a by-product of coordination?''.
The answer is ``No'', since we observe agent populations that converge to using successful misunderstandings, i.e. high levels of coordination using misaligned interpretations, at the end of phase 1 in both experiments 1 (Sect. \ref{subsec:pop_increase_exp_2_to_3}) and 3 (Sec. \ref{subsec:pairing_restrictions_exp}).
To answer RQ2, ``When do misaligned interpretations cause coordination failure?'', we study the robustness of successful misunderstandings.
%
%
%
Agent populations that coordinated using successful misunderstandings were not able to learn to coordinate with new interaction partners, as seen in phase 2 of both experiments 1 (Sec. \ref{subsec:pop_increase_exp_2_to_3}) and 3 (Sec. \ref{subsec:pairing_restrictions_exp}).
%
%
Finally, to answer RQ3, i.e. ``What conditions ensure the emergence of shared interpretations?'', we conclude in Sec. \ref{subsec:necessary_conditions_summary} that populations need to have a group of at least 3 agents that interact with each other in all possible scenarios, i.e. environment states. This groups ensures the emergence of shared interpretations.
%
%
%
%
%
%
%
The presented experiments are validated in environments that make coordination establishment more difficult, as presented in Sec. \ref{subsec:validation_on_more_complex_reward_function}.

Future directions of this study can repeat experiments while including humans in agent populations.
%
%
Moreover, identifying experimental setups that require unrestricted agent populations of 4 agents, instead of 3, to avoid successful misunderstandings, can lead towards developing a broader theory.
%
%
%
%
Last but not least, future studies can focus on identifying possible misalignment and its consequences between humans and large language models.

\vspace{-0.1cm}

\begin{credits}
\subsubsection{\ackname}
This work was supported by ``MUHAI - Meaning and Understanding in Human-centric Artificial Intelligence'' project, funded by the European Union's Horizon 2020 research and innovation program under grant agreement No 951846.
\begin{minipage}{0.05\textwidth}
  \includegraphics[trim={0cm 0cm 0cm 0cm},width=\textwidth]{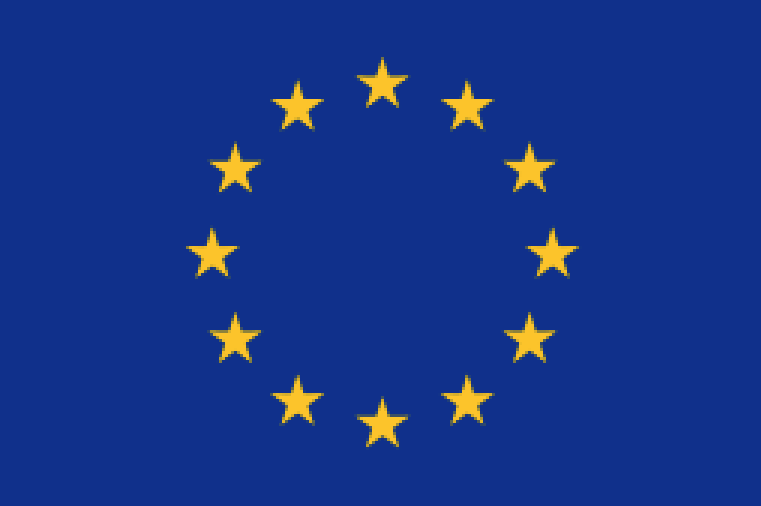}
\end{minipage}%

\noindent

\end{credits}

\newpage


\bibliographystyle{splncs04} 

\bibliography{bibliography}

\begin{thebibliography}{10}
\providecommand{\url}[1]{\texttt{#1}}
\providecommand{\urlprefix}{URL }
\providecommand{\doi}[1]{https://doi.org/#1}

\bibitem{aberer2003chatty}
Aberer, K., Cudr{\'e}-Mauroux, P., Hauswirth, M.: The chatty web: Emergent semantics through gossiping. In: Proceedings of the 12th international conference on World Wide Web. pp. 197--206 (2003)

\bibitem{anslow}
Anslow, M., Rovatsos, M.: Aligning experientially grounded ontologies using language games. In: Croitoru, M., Marquis, P., Rudolph, S., Stapleton, G. (eds.) Graph Structures for Knowledge Representation and Reasoning. pp. 15--31. Springer International Publishing, Cham (2015)

\bibitem{atencia_automata}
Atencia, M., Schorlemmer, M.: An interaction-based approach to semantic alignment. Web Semantics: Science, Services and Agents on the World Wide Web  \textbf{s 12–13},  131–147 (04 2012). \doi{10.1016/j.websem.2011.12.001}

\bibitem{bender_koller_octapus_test}
Bender, E.M., Koller, A.: Climbing towards {NLU}: {On} meaning, form, and understanding in the age of data. In: Jurafsky, D., Chai, J., Schluter, N., Tetreault, J. (eds.) Proceedings of the 58th Annual Meeting of the Association for Computational Linguistics. pp. 5185--5198. Association for Computational Linguistics, Online (Jul 2020). \doi{10.18653/v1/2020.acl-main.463}, \url{https://aclanthology.org/2020.acl-main.463/}

\bibitem{emergent_semantics}
Cudr{\'e}-Mauroux, P.: Emergent Semantics, pp. 982--985. Springer US, Boston, MA (2009). \doi{10.1007/978-0-387-39940-9_1311}, \url{https://doi.org/10.1007/978-0-387-39940-9\_1311}

\bibitem{jerome}
Euzenat, J.: First experiments in cultural alignment repair (extended version). In: The Semantic Web: ESWC 2014 Satellite Events. p. 115–130. Springer-Verlag, Berlin, Heidelberg (2014). \doi{10.1007/978-3-319-11955-7_10}, \url{https://doi-org.vu-nl.idm.oclc.org/10.1007/978-3-319-11955-7\_10}

\bibitem{learning_to_communicate_forester}
Foerster, J., Assael, I.A., de~Freitas, N., Whiteson, S.: Learning to communicate with deep multi-agent reinforcement learning. In: Lee, D., Sugiyama, M., Luxburg, U., Guyon, I., Garnett, R. (eds.) Advances in Neural Information Processing Systems. vol.~29. Curran Associates, Inc. (2016)

\bibitem{human_communication_emergence}
Galantucci, B.: An experimental study of the emergence of human communication systems. Cognitive science  \textbf{29},  737--67 (09 2005). \doi{10.1207/s15516709cog0000_34}

\bibitem{shortcut_learning}
Geirhos, R., Jacobsen, J., Michaelis, C., Zemel, R.S., Brendel, W., Bethge, M., Wichmann, F.A.: Shortcut learning in deep neural networks. CoRR  \textbf{abs/2004.07780} (2020), \url{https://arxiv.org/abs/2004.07780}

\bibitem{shared_understanding_framework}
Kondylidis, N., Tiddi, I., ten Teije, A.: A framework for establishing shared, task-oriented understanding in hybrid open multi-agent systems. Frontiers in Artificial Intelligence  \textbf{Volume 8 - 2025} (2025). \doi{10.3389/frai.2025.1440582}, \url{https://www.frontiersin.org/journals/artificial-intelligence/articles/10.3389/frai.\\2025.1440582}

\bibitem{lazaridou}
Lazaridou, A., Peysakhovich, A., Baroni, M.: Multi-agent cooperation and the emergence of (natural) language. CoRR  \textbf{abs/1612.07182} (2016), \url{http://arxiv.org/abs/1612.07182}

\bibitem{Lewis1970ConventionAP}
Lewis, D.K.: Convention: A Philosophical Study. Wiley-Blackwell, Cambridge, MA, USA (1969)

\bibitem{human_agent_study}
Mohan, S., Laird, J.: Learning goal-oriented hierarchical tasks from situated interactive instruction. Proceedings of the AAAI Conference on Artificial Intelligence  \textbf{28} (06 2014). \doi{10.1609/aaai.v28i1.8756}

\bibitem{nevens2019interactive}
Nevens, J., Van~Eecke, P., Beuls, K.: Interactive learning of grounded concepts. In: BNAIC/BENELEARN (2019)

\bibitem{psychology_of_dialogue}
Pickering, M.J., Garrod, S.: Toward a mechanistic psychology of dialogue. Behavioral and Brain Sciences  \textbf{27}(2),  169–190 (2004). \doi{10.1017/S0140525X04000056}

\bibitem{rita2022emergentcommunicationgeneralizationoverfitting}
Rita, M., Tallec, C., Michel, P., Grill, J.B., Pietquin, O., Dupoux, E., Strub, F.: Emergent communication: Generalization and overfitting in lewis games (2022), \url{https://arxiv.org/abs/2209.15342}

\bibitem{russell2019human_compatible}
Russell, S.: Human compatible: AI and the problem of control. Penguin Uk (2019)

\bibitem{Searle_1980_chinese_room}
Searle, J.R.: Minds, brains, and programs. Behavioral and Brain Sciences  \textbf{3}(3),  417–424 (1980). \doi{10.1017/S0140525X00005756}

\bibitem{steels_language_2012}
Steels, L.: Experiments in Cultural Language Evolution. John Benjamins (2012). \doi{10.1075/ais.3}, luc Steels and Martin Loetzsch (2012). The Grounded Naming Game. p. 41-59

\end{thebibliography}

\end{document}